\documentclass[11pt]{article}

\usepackage[final]{acl}
\usepackage{enumitem}
\usepackage{algorithm}
\usepackage{algorithmic}
\usepackage{times}
\usepackage{latexsym}
\usepackage{multirow}
\usepackage[T1]{fontenc}
\usepackage{amssymb}
\usepackage{subcaption}
\usepackage{tabularx}
\usepackage[utf8]{inputenc}
\usepackage{amsmath}
\usepackage{microtype}
\usepackage{booktabs}
\usepackage{inconsolata}
\usepackage{hyperref}

\usepackage{graphicx}
\usepackage{framed}
\usepackage[breakable]{tcolorbox}

\title{Beyond Isolated Behaviors: Hierarchical User Modeling for LLM Personalization}

\author{
  Liang Wang\textsuperscript{1*,3},
  Xinyi Mou\textsuperscript{1*},
  Xiaoyou Liu\textsuperscript{1,3},
  Tiannan Wang\textsuperscript{3},
  Yuqing Wang\textsuperscript{3},
  Zhongyu Wei\textsuperscript{1,2\textdagger} \\
  \textsuperscript{1}School of Data Science, Fudan University \\
  \textsuperscript{2}Shanghai Innovation Institute \\
  \textsuperscript{3}OPPO \\
  \texttt{liangwang25,xiaoyouliu25@m.fudan.edu.cn, xymou20,zywei@fudan.edu.cn}
}

\begin{document}
\maketitle

\begin{abstract}

Large Language Models (LLMs) have demonstrated remarkable capabilities across diverse domains, yet personalizing their outputs to individual users remains an open challenge. Existing approaches predominantly adopt a flat behavioral paradigm, aggregating user behaviors without an explicit account of how they are organized into deeper behavioral structures. In this work, we draw on Pierre Bourdieu's Theory of Practice to propose PHF (Practice-Habitus-Field), a sociologically grounded framework that reconceptualizes LLM personalization through three hierarchical levels: individual behaviors as practices, their temporal accumulation into stable dispositions as habitus, and shared regularities across similar users as fields. We instantiate PHF through $\mathrm{PHF}_{\text{Compass}}$, a lightweight and model-agnostic implementation based on a frozen LLM. Experiments on the Language Model Personalization (LaMP) benchmark demonstrate consistent improvements across diverse tasks, while further analyses validate the interpretability and extensibility of the learned behavioral structures.\footnote{Anonymous code repository: \url{https://anonymous.4open.science/r/PHF-0123}. Work completed during the internship of Liang Wang and Xiaoyou Liu at OPPO.}

\end{abstract}

\section{Introduction}

Large language models (LLMs) have demonstrated remarkable capabilities across diverse tasks and domains, driving growing interest in personalized language generation~\cite{he2025simulation,deng2023plug,ning2025user,hebert2024persoma,deng2025onerec,li2024agent,bao2024piors}. In personalized settings, user expectations are inherently heterogeneous: responses appropriate for one user may not align with the expectations of another. Consequently, effective personalization requires models to capture the underlying behavioral patterns that shape user preferences and behaviors.

\begin{figure}
    \centering
    \includegraphics[width=1\linewidth]{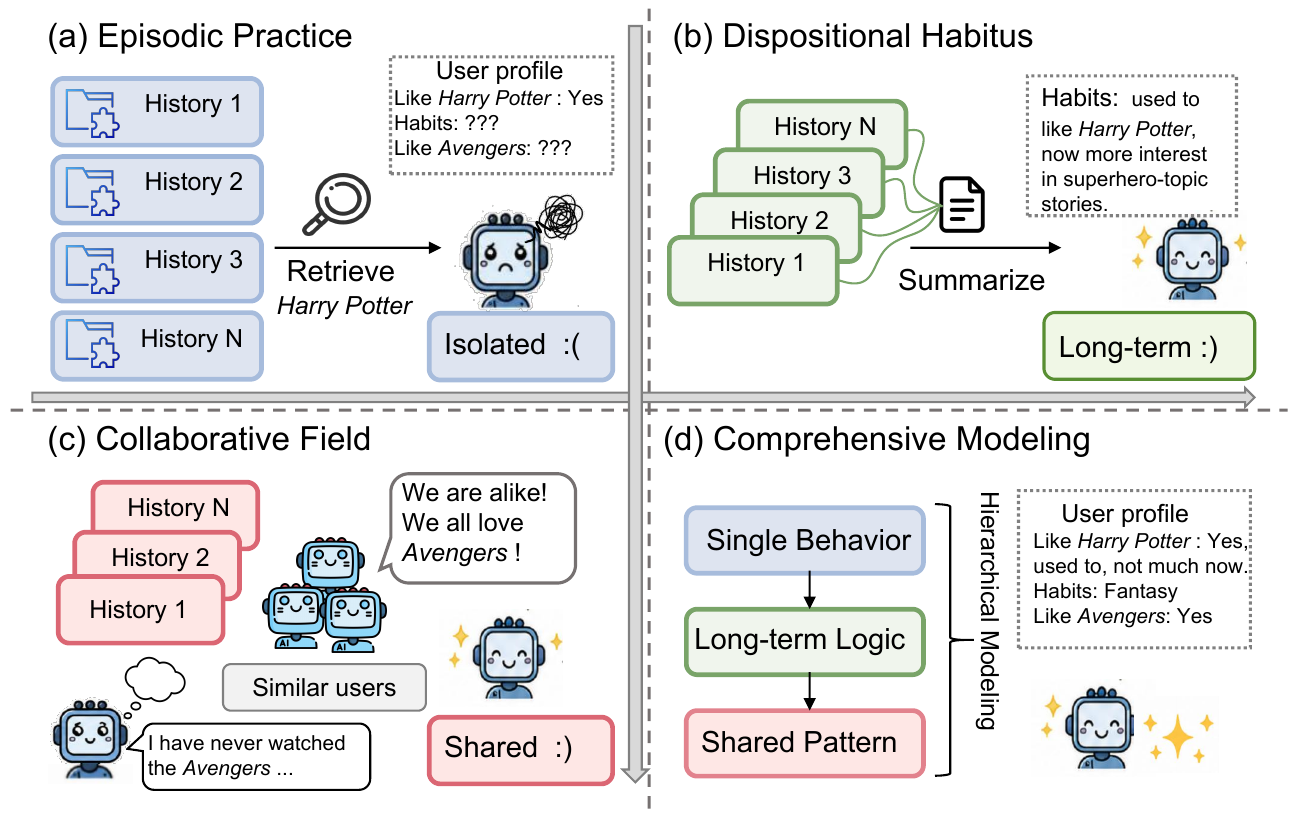}
    \caption{Comparison of personalization paradigms. (a) Existing methods adopt a flat behavioral paradigm, retrieving or encoding user behaviors without explicit structural modeling. (b) Habitus captures the temporal accumulation of repeated behaviors into stable dispositions. (c) Field captures shared behavioral regularities among users with similar dispositions. (d) PHF integrates practice, habitus, and field into a unified hierarchical framework for LLM personalization.}
    \label{fig:intro}
    \vspace{-10pt} 
\end{figure}

Recent approaches to LLM personalization have explored retrieval-augmented prompting, learned user representations, and parameter-efficient adaptation~\cite{salemi2024optimization,tan-etal-2024-democratizing,liu2024llms+}. While these methods have shown promising results, they predominantly adopt a \emph{flat} behavioral paradigm that treats user behaviors as an unordered collection of observed instances, as illustrated in Figure.~\ref{fig:intro}(a). This flat formulation gives rise to two key limitations: (1) \textbf{behavior fragmentation}: within each user, behavioral histories lack internal structure, where they are treated as isolated instances rather than being consolidated into stable long-term patterns. (2) \textbf{user isolation}: users are modeled independently, without leveraging shared behavioral regularities across users with similar behavioral profiles, limiting generalization when user history is sparse. Together, these limitations make it difficult to capture the multi-level behavioral structures that govern user behavior, leading to suboptimal personalization.



Addressing these limitations requires a structured view of user behavior, modeling both how behaviors consolidate within individuals and how behavioral regularities are shared across users. We draw inspiration from Pierre Bourdieu's Theory of Practice~\cite{bourdieu1990logic,bourdieu2018distinction,bourdieu2020outline}, a well-established sociological framework that organizes human behavior around repeated actions, durable dispositions, and shared social spaces. Based on this perspective, we propose \textbf{PHF} (Practice-Habitus-Field), a hierarchical user modeling framework that addresses both limitations through a unified multi-level representation. In PHF, individual behaviors are first abstracted as \textit{practices}, which progressively consolidate into \textit{habitus}, stable behavioral dispositions that capture the long-term tendencies within each user (Figure.~\ref{fig:intro}(b)). Users with similar habitus are further organized into shared \textit{fields}, which encode collective behavioral regularities (Figure.~\ref{fig:intro}(c)). Collectively, PHF formulates LLM personalization as a hierarchical behavioral modeling problem (Figure.~\ref{fig:intro}(d)) that jointly captures within-user disposition formation and cross-user behavioral structure.

To implement PHF, we develop $\mathrm{PHF}_{\text{Compass}}$ (\underline{CO}mbined \underline{M}odeling of \underline{P}ractice, \underline{A}ccumulated \underline{S}tructure, and \underline{S}hared signals), a lightweight model compatible with diverse backbones. $\mathrm{PHF}_{\text{Compass}}$ represents practices by mapping individual behaviors into discrete semantic prototypes, filtering surface-level noise while preserving core behavioral semantics. 
These practices are then temporally aggregated into habitus representations that encode stable long-term behavioral dispositions for each user. To capture collective structure, $\mathrm{PHF}_{\text{Compass}}$ further organizes users into latent fields derived from their habitus, enabling users with similar dispositions to benefit from shared behavioral regularities. During inference, both habitus and field representations are provided to the LLM. This enables personalized generation that reflects both individualized dispositions and broader collective patterns. Our main contributions are summarized as follows:

(1) We propose \textbf{PHF} (Practice-Habitus-Field), a sociologically grounded framework for LLM personalization that draws on Bourdieu's Theory of Practice. PHF introduces a hierarchical behavioral structure comprising practice, habitus, and field, providing a unified perspective that captures both the temporal coherence within individual behavior and the shared regularities across behaviorally similar users.

(2) We develop \textbf{$\mathrm{PHF}_{\text{Compass}}$}, a concrete instantiation of PHF that operationalizes the three constructs through residual behavioral abstraction, temporal aggregation, and collaborative clustering. $\mathrm{PHF}_{\text{Compass}}$ is lightweight, model-agnostic, and requires no modification to the underlying LLM.

(3) Extensive experiments on the Language Model Personalization (LaMP) benchmark demonstrate that $\mathrm{PHF}_{\text{Compass}}$ achieves consistent improvements across both classification and generation tasks. Further analyses validate the interpretability of the learned behavioral structures and support the utility of the proposed sociological perspective.


\begin{figure*}[t]
    \centering
    \includegraphics[width=\linewidth]{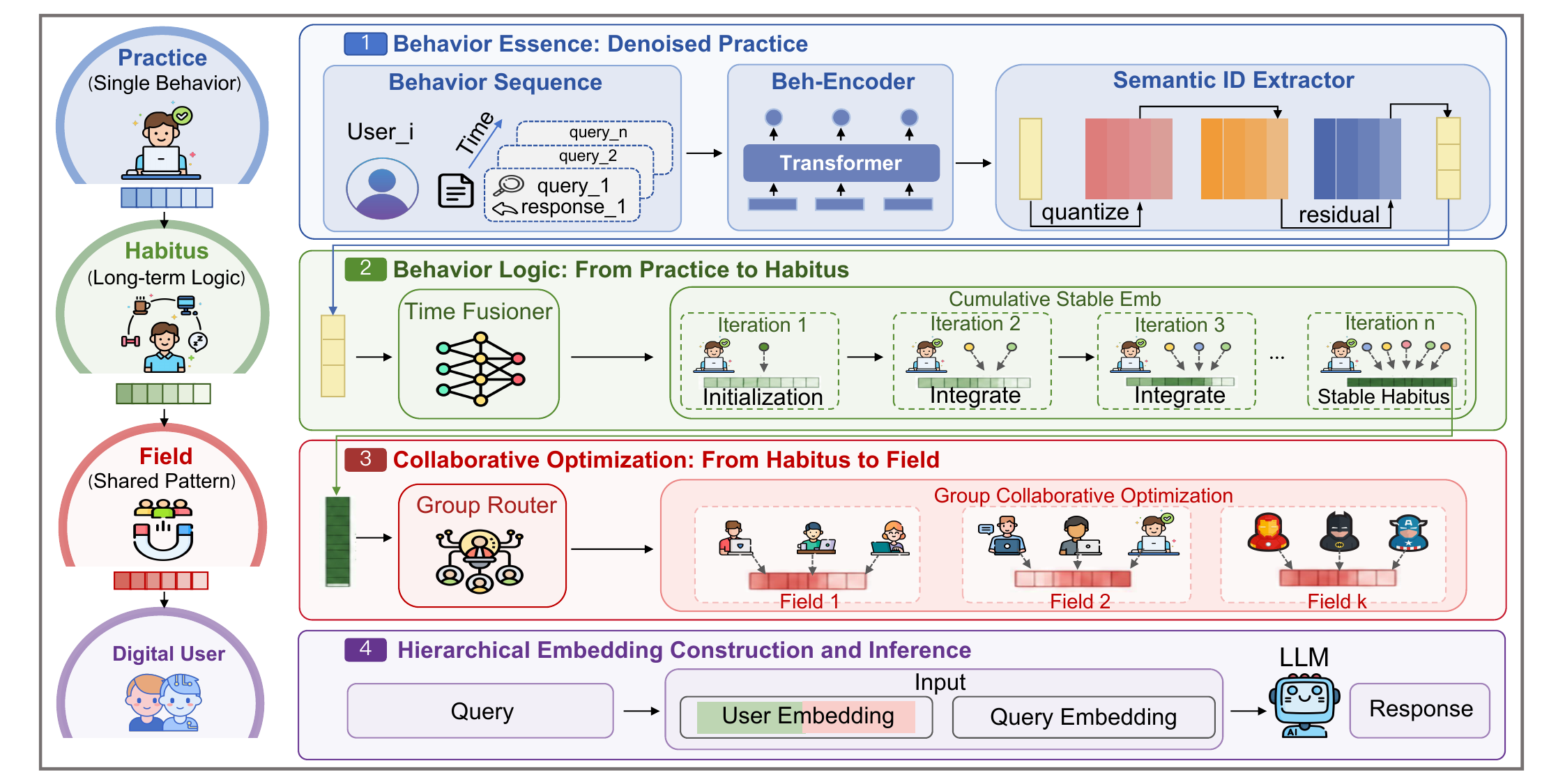}
    \caption{ Overview of the PHF framework and $\mathrm{PHF}_{\text{Compass}}$ implement. (1) Raw behaviors are abstracted into denoised practice essence via semantic ID extraction. (2) Practices are temporally aggregated  into stable habitus representations. (3) Users are organized into latent fields through group router. (4) Habitus and field embeddings are combined as well as the query embedding to condition the LLM for personalized generation.}
    \label{fig:framework}
\end{figure*}

\section{Related Work}

\subsection{LLM-based Personalization}

Despite strong general capabilities in reasoning, knowledge-intensive tasks, and domain applications~\cite{joshi2017triviaqa,liang2025hardcorelogic,wu2025finteam,jia2025ready}, LLMs remain fundamentally limited in personalization. Since standard LLMs are trained on broadly aggregated corpora, they capture general linguistic and task patterns but contain limited information about individual behavioral tendencies, preferences, and long-term decision logic~\cite{tao2025personafeedback,xi2025rise}.
Existing LLM personalization methods can be broadly characterized by how they utilize user behavioral histories. Retrieval-based methods personalize generation by selecting relevant behaviors from user history and incorporating them into prompts~\cite{salemi2023lamp,kumar2024longlamp,shi2025retrieval,zhang2024personalized,wang2025mem}. While flexible, these approaches operate at the level of individual behaviors without aggregating them into a holistic user representation. Aggregation-based methods instead construct unified user representations through summarized preferences~\cite{zhang2024guided}, learned soft prompts or PEFT modules~\cite{zhuang2024hydra,zhang2025proper}, compressed user embeddings~\cite{liu2024llms+}, or retrieval optimization~\cite{salemi2024optimization}. These methods merge behavioral signals into user-level representations but do so without explicitly modeling how behaviors are internally organized or how regularities are shared across users. In both cases, user behavior is treated as a flat collection of observed instances, leaving the within-user behavioral structure and cross-user behavioral regularity largely unaddressed.

\subsection{User Embedding Modeling}

Learning user representations from behavioral data has long been central to personalized systems~\cite{elkahky2015multi,purificato2024user,ko2022survey,mu2018survey}. Recent LLM personalization methods further model users through textual profiles, behavioral summaries, or trainable embeddings derived from interaction histories~\cite{hebert2024persoma,doddapaneni2024user,liu2024llms+}. Temporal weighting and semantic quantization techniques have also been introduced to better capture salient behavioral signals~\cite{zare2025tillm,dai2022spatio,rajput2023recommender,zheng2024adapting,deng2025onerec,chen2025onesearch}. CURP~\cite{wang2026curp} models users with both a stable inner representation and a query-aware dynamic representation via PQ quantization, balancing the effectiveness and efficiency. Different from prior work that primarily models behaviors independently, our approach organizes user behaviors into hierarchical behavioral structures, capturing both stable individual dispositions and collective regularities across users.

\section{Method}




\subsection{PHF Framework Setup}

We introduce PHF, a hierarchical user modeling framework for LLM personalization, illustrated in the left part of Figure.~\ref{fig:framework}. PHF organizes personalization through three behavioral levels: \emph{practice}, \emph{habitus}, and \emph{field}. Individual user behaviors are first treated as practices, which are progressively accumulated into habitus, stable behavioral dispositions that characterize each user's long-term tendencies. Users with similar habitus are further organized into shared fields that capture collective behavioral regularities across behaviorally similar users. Personalized generation is conditioned on both habitus and field representations, yielding the following task formulation:

Let $\mathcal{U} = \{u_1, u_2, \dots, u_N\}$ denote a set of $N$ users. Each user $u_i$ is associated with a chronologically ordered interaction history $\mathcal{H}_i = \{h_{i1}, h_{i2}, \dots, h_{iT_i}\}$, where each $h_{it} = (q_{it}, r_{it})$ is a query-response pair. Given a new query $\mathcal{Q}_i$, the goal is to generate a personalized response conditioned on the user's behavioral history: $P(\mathcal{R}_i \mid \mathcal{Q}_i, \mathcal{H}_i)$. Following the PHF hierarchy, each interaction is abstracted into a practice representation, temporally aggregated into a habitus embedding $\mathbf{h}_i$, and associated with a field embedding $\mathbf{f}_i$. Personalized generation is thus conditioned on both individual and collective signals:
\begin{equation}
P(\mathcal{R}_i \mid \mathcal{Q}_i, \mathbf{h}_i, \mathbf{f}_i).
\end{equation}




\subsection{$\mathrm{PHF}_{\text{Compass}}$ Architecture}

To instantiate the PHF framework, we develop $\mathrm{PHF}_{\text{Compass}}$, a lightweight and model-agnostic architecture that operationalizes the three levels of PHF through dedicated model components. The architecture is shown in the right part of Figure.~\ref{fig:framework} and training consists of two stages. In the first stage, we train the model to learn improved \textit{practice} representations. In the second stage, these optimized representations are used to further construct \textit{habitus} and \textit{field}, which are then jointly leveraged for personalized generation.


\paragraph{Enhanced Practice: Denoised Behavior Essence}

As illustrated in Figure.~\ref{fig:framework}-(1), raw user behaviors
often contain substantial surface variation. Directly modeling continuous embeddings risks overfitting to contextual details rather than capturing stable behavioral semantics. To obtain more robust representations, we follow recent semantic quantization approaches~\cite{deng2025onerec,chen2025onesearch} and abstract each interaction into a discrete semantic prototype~\cite{wang2026curp}. This discretization serves as a behavioral abstraction process: behaviors with different surface forms but similar underlying intent converge toward shared prototypes.

Given an interaction $h_{it}$, we first encode it into a dense representation:
\begin{equation}
    \mathbf{x}_{it} = \mathrm{Enc}(h_{it}),
\end{equation}
where $\mathrm{Enc}(\cdot)$ denotes a lightweight encoder. Detailed implementation choices are deferred to the Appendix~\ref{sec:CL}.

We then apply multi-level residual vector quantization to progressively decompose each interaction into coarse-to-fine semantic components:
\begin{equation}
    \mathbf{q}_l = \arg\min_{\mathbf{c}\in\mathcal{C}_l} \| \mathbf{r}_{l-1}-\mathbf{c} \|_2, \qquad \mathbf{r}_l = \mathbf{r}_{l-1}-\mathbf{q}_l,
\end{equation}
where $\mathbf{r}_0=\mathbf{x}_{it}$ and $\mathcal{C}_l$ denotes the codebook at the $l$-th quantization level. Higher levels capture dominant behavioral intent, while lower levels progressively refine finer-grained distinctions. The final practice representation is reconstructed as:
\begin{equation}
    \hat{\mathbf{x}}_{it} = \sum_{l=1}^{L}\mathbf{q}_l,\mathrm{SID}(p_{it}) = (i_1,i_2,\dots,i_L),
\end{equation}
where $(i_1,\dots,i_L)$ denotes the hierarchical Semantic ID of the interaction.

The quantization module is optimized through $\mathcal{L}_{\text{RQ}}$, a combination of reconstruction, commitment, and codebook regularization objectives:
\begin{equation}
\begin{split}
    \mathcal{L}_{\text{RQ}} = \| \mathbf{x}_{it}-\hat{\mathbf{x}}_{it} \|_2^2 
    + \lambda_v \sum_{l=1}^{L} \| \mathbf{r}_{l-1}-\mathrm{sg}(\mathbf{q}_l) \|_2^2 \\
    + \lambda_v \sum_{l=1}^{L} \| \mathrm{sg}(\mathbf{r}_{l-1})-\mathbf{q}_l \|_2^2 
    + \lambda_u \sum_{l=1}^{L} \mathrm{KL}( \bar{p}_l \| \mathcal{U} ).
\end{split}
\end{equation}
The first term minimizes reconstruction error between the original and quantized representations. The second and third terms are symmetric commitment losses that encourage the encoder outputs and codebook entries to converge, where $\mathrm{sg}(\cdot)$ denotes the stop-gradient operator. The final term is a codebook utilization regularizer that penalizes deviation of the codeword usage distribution $\bar{p}_l$ from the uniform distribution $\mathcal{U}$, preventing codebook collapse. Following prior work~\cite{deng2025onerec,chen2025onesearch}, the codebooks are initialized through balanced K-Means and are maintained as task-specific. Additional implementation details are provided in the Appendix~\ref{sec:bkl}.



\paragraph{Habitus: Modeling Long-Term Behavioral Dispositions}

As illustrated in Figure.~\ref{fig:framework}-(2), to model the stable behavioral dispositions formed through repeated behaviors, we aggregate the sequence of denoised practice representations $\{\hat{\mathbf{x}}_{it}\}_{t=1}^{T_i}$ for user $u_i$ over time with a time fusioner. Since more recent behaviors tend to better reflect a user's current preferences, we simplify this as a temporally weighted aggregation:
\begin{equation}
\mathbf{h}_i
=
\sum_{t=1}^{T_i}
w_t
\hat{\mathbf{x}}_{it},
\qquad
w_t
=
\frac{
\exp(\sqrt{t})
}{
\sum_{\tau=1}^{T_i}
\exp(\sqrt{\tau})
}.
\end{equation}
The square-root scaling ensures that the recency effect grows sublinearly, preventing recent behaviors from dominating the representation while still preserving temporal sensitivity. The resulting habitus vector $\mathbf{h}_i$ encodes a coherent representation of the user's long-term behavioral disposition, which subsequently serves as the basis for modeling collective behavioral structure in the field.


\paragraph{Field: Collective Behavioral Structure}

As illustrated in Figure.~\ref{fig:framework}-(3), to capture shared behavioral regularities across users, we organize users into latent fields through a group router based on their habitus representations. The underlying intuition is that users with similar long-term behavioral dispositions tend to exhibit similar preferences, suggesting that shared behaviorals can complement purely individual modeling. This idea is related in spirit to collaborative filtering~\cite{shi2025retrieval}
, where similar users provide useful inductive bias for one another. However, unlike classical collaborative filtering, personalization datasets typically lack sufficient overlapping observations across identical items or contexts. We therefore approximate this shared structure by inducing latent fields through clustering over user-level habitus representations.
 

Specifically, we induce latent fields by partitioning users with their habitus via K-Means:
\begin{equation}
\{
\boldsymbol{\mu}_k
\}_{k=1}^{K}
=
\arg\min_{\{\boldsymbol{\mu}_k\}}
\sum_{i=1}^{N}
\min_k
\|
\mathbf{h}_i-\boldsymbol{\mu}_k
\|_2^2,
\end{equation}
where each centroid $\boldsymbol{\mu}_k$ corresponds to a field embedding that encodes the shared behavioral disposition of users within that cluster. Each user is then assigned to the nearest field:
\begin{equation}
\mathbf{f}_i
=
\boldsymbol{\mu}_{k^*},
\qquad
k^*
=
\arg\min_k
\|
\mathbf{h}_i-\boldsymbol{\mu}_k
\|_2.
\end{equation}
The resulting field representations capture collective behavioral regularities across users with similar long-term dispositions, providing complementary signals that are particularly informative under sparse-history and cold-start settings.

\subsection{Generation with Hierarchical User Representation}

Finally, as shown in Figure.~\ref{fig:framework}-(4), we condition the downstream LLM on both habitus and field representations. Specifically, we project them into the LLM embedding space:
\begin{equation}
\phi_h :
\mathbb{R}^d
\rightarrow
\mathbb{R}^{d_{\text{llm}}},
\qquad
\phi_f :
\mathbb{R}^d
\rightarrow
\mathbb{R}^{d_{\text{llm}}}.
\end{equation}

The final input sequence is constructed as:
\begin{equation}
\mathbf{X}_{\text{in}}
=
\left[
\phi_h(\mathbf{h}_i),
\,
\phi_f(\mathbf{f}_i),
\,
\mathrm{Emb}(\mathcal{Q}_i)
\right].
\end{equation}

The model is optimized using the standard language modeling objective:
\begin{equation}
\mathcal{L}_{\text{LM}}
=
-
\sum_{t=r}^{M}
\log
P(
x_t
\mid
\mathbf{X}_{\text{in}},
x_{<t}
).
\end{equation}

Only lightweight projection layers are trained, while the underlying LLM remains frozen. This design enables scalable personalization grounded in both individual behavioral dispositions and collective social structure. During inference, each user is assigned to the nearest field based on cluster centroids learned during training, requiring no access to other users' data.

\begin{table*}[t]
\centering
\resizebox{\textwidth}{!}{%
\begin{tabular}{ccccccccccccccccccccccc}
\hline
\multirow{2}{*}{\textbf{Task}} & \multirow{2}{*}{\textbf{Metric}} &  & \multirow{2}{*}{\begin{tabular}{c}\textbf{Zero}\\\textbf{Shot}\end{tabular}} &  & \multicolumn{4}{c}{\textbf{Flat: Retrieval-Based}} &  & \multicolumn{3}{c}{\textbf{Flat: Aggregation-Based}} &  & \multicolumn{6}{c}{\textbf{PHF Implements}} \\ \cline{6-9} \cline{11-13} \cline{15-20}
 &  &  &  &  & Recency & BM25 & Dense & ROPG &  & OPPU & DiffMean & PPlug &  & T-H & T-F & L-F & T-H+T-F & T-H+L-F & $\mathrm{PHF}_{\text{Compass}}$ \\ \hline
LaMP-1 & Acc$\uparrow$ &  & 0.464 &  & 0.504 & 0.504 & 0.529 & 0.526 &  & - & - & 0.520 &  & 0.488 & 0.487 & 0.535 & 0.525 & \underline{0.569} & \textbf{0.631} \\ \hline
\multirow{2}{*}{LaMP-2} & Acc$\uparrow$ &  & 0.366 &  & 0.449 & 0.453 & 0.490 & 0.489 &  & 0.444 & 0.374 & 0.503 &  & 0.457 & 0.467 & \underline{0.528} & 0.518 & 0.506 & \textbf{0.531} \\
 & F1$\uparrow$ &  & 0.337 &  & 0.432 & 0.437 & 0.468 & 0.455 &  & 0.400 & 0.343 & 0.448 &  & 0.444 & 0.433 & \underline{0.490} & \textbf{0.478} & 0.474 & 0.461 \\ \hline
\multirow{2}{*}{LaMP-3} & MAE$\downarrow$ &  & 0.498 &  & 0.456 & 0.495 & 0.414 & 0.483 &  & 0.426 & 0.504 & \underline{0.314} &  & 0.355 & 0.367 & 0.419 & 0.324 & 0.334 & \textbf{0.260} \\
 & RMSE$\downarrow$ &  & 0.825 &  & 0.811 & 0.859 & 0.750 & 0.840 &  & 0.740 & 0.834 & 0.805 &  & 0.691 & 0.725 & 0.762 & 0.660 & \underline{0.655} & \textbf{0.579} \\ \hline
\multirow{2}{*}{LaMP-4} & R-1$\uparrow$ &  & 0.145 &  & 0.167 & 0.170 & 0.171 & 0.165 &  & \underline{0.192} & 0.146 & 0.182 &  & 0.168 & 0.152 & 0.168 & 0.158 & 0.174 & \textbf{0.193} \\
 & R-L$\uparrow$ &  & 0.123 &  & 0.150 & 0.151 & 0.153 & 0.148 &  & \underline{0.170} & 0.126 & 0.162 &  & 0.151 & 0.136 & 0.151 & 0.140 & 0.155 & \textbf{0.172} \\ \hline
\multirow{2}{*}{LaMP-5} & R-1$\uparrow$ &  & 0.441 &  & 0.461 & 0.463 & 0.465 & 0.461 &  & 0.475 & 0.442 & \underline{0.512} &  & 0.471 & 0.476 & 0.476 & 0.485 & 0.489 & \textbf{0.516} \\
 & R-L$\uparrow$ &  & 0.360 &  & 0.387 & 0.389 & 0.391 & 0.385 &  & 0.398 & 0.358 & \underline{0.443} &  & 0.392 & 0.398 & 0.403 & 0.411 & 0.412 & \textbf{0.453} \\ \hline
\multirow{2}{*}{LaMP-7} & R-1$\uparrow$ &  & 0.448 &  & 0.418 & 0.464 & 0.462 & 0.447 &  & - & 0.444 & \underline{0.524} &  & 0.456 & 0.414 & 0.463 & 0.418 & 0.459 & \textbf{0.535} \\
 & R-L$\uparrow$ &  & 0.392 &  & 0.365 & 0.408 & 0.406 & 0.393 &  & - & 0.388 & \underline{0.465} &  & 0.400 & 0.365 & 0.406 & 0.369 & 0.403 & \textbf{0.475} \\ \hline
\end{tabular}
}
\caption{Performance of all methods on the LaMP benchmark. Existing personalization methods are organized under the flat paradigm, while PHF Implements validate each level of the proposed hierarchy. The best results are shown in \textbf{bold}, and the second-best are \underline{underlined}. $\uparrow$ indicates that higher values are better. ``--'' denotes that the method is not applicable. All metrics are the average of 3 independent reruns with different random seeds.}
\label{tab:main results}
\vspace{-10pt} 
\end{table*}

\section{Experiments}

\subsection{Experiment Setup}
\subsubsection{Dataset}
We conduct experiments on the Language Model Personalization (LaMP) benchmark~\cite{salemi2023lamp}. Following prior work~\cite{liu2024llms+,shi2025retrieval,tan-etal-2024-democratizing,tan-etal-2024-personalized,zhang2025steer}, we evaluate our framework on all six public tasks\footnote{We didn't include LaMP-6 Personalized Email Subject Generation as it is not publicly available.}. The selected tasks span both 3 classification tasks and 3 generation tasks. The classification tasks contain: 
(1) \textbf{LaMP-1}: Personalized Citation Identification; 
(2) \textbf{LaMP-2}: Personalized Movie Tagging; 
(3) \textbf{LaMP-3}: Personalized Product Rating; 
The generation tasks contain: 
(4) \textbf{LaMP-4}: Personalized News Headline Generation;  
(5) \textbf{LaMP-5}: Personalized Scholarly Title Generation; and  
(6) \textbf{LaMP-7}: Personalized Tweet Paraphrasing.  
Since the test sets are held out by the benchmark organizers, we report results on the official validation sets using the provided time-based splits. More details are provided in Appendix~\ref{sec:data}.

\subsubsection{Evaluation Metrics}
Following the default evaluation protocol of the LaMP benchmark, we adopt task-specific metrics as follows:  
\textbf{Accuracy} for LaMP-1;  
\textbf{Accuracy} and \textbf{F1-score} for LaMP-2;  
\textbf{Mean Absolute Error (MAE)} and \textbf{Root Mean Squared Error (RMSE)} for LaMP-3;  
and \textbf{ROUGE-1 (R-1)}, \textbf{ROUGE-L (R-L)} for LaMP-4, LaMP-5, and LaMP-7.

\subsection{Implementation Details}

We use \textbf{Contriever}~\cite{izacard2021unsupervised} as the default encoder and \textbf{Qwen2.5-7B-Instruct}~\cite{qwen2.5} as the decoder. 
For residual quantization codebooks, we use $L = 3$ layers with codebook sizes of $32$, $32$, and $16$. The default number of retrieved histories is set to $k = 8$, and the number of field clusters is set to $K = 10$. We optimize the model using AdamW with a learning rate of $1 \times 10^{-4}$. All experiments are conducted on 8 NVIDIA A100 (80GB) GPUs. Additional details are provided in the Appendix~\ref{sec:detail}.

\subsection{Baselines}

We compare $\mathrm{PHF}_{\text{Compass}}$ against representative personalization methods organized in 3 categories, along with alternative implements of PHF.

\textbf{Zero-Shot}:
The base \textbf{Qwen2.5-7B-Instruct} 
model generates responses solely from the input query without incorporating any user information.

\textbf{Flat: Retrieval-Based}:
These methods personalize by selecting individual behaviors from user history without aggregating them into a unified representation.
(1) \textbf{Recency} retrieves the most recent $k$ behaviors according to temporal order.
(2) \textbf{BM25}~\cite{bm25} retrieves query-relevant histories using sparse lexical matching.
(3) \textbf{Dense}~\cite{izacard2021unsupervised} uses dense semantic embeddings for retrieval.
(4) \textbf{ROPG}~\cite{salemi2024optimization} further improves dense retrieval through reinforcement learning optimization of the retriever.

\textbf{Flat Aggregation-Based}:
These methods construct a unified user representation by aggregating behaviors, but without explicitly modeling how behaviors are structured.
(1) \textbf{OPPU}~\cite{tan-etal-2024-democratizing} trains dedicated LoRA~\cite{hu2022lora} adapters for each user.
(2) \textbf{DiffMean}~\cite{zhang2025steer} models user style through residual activation differences between personalized and non-personalized responses.
(3) \textbf{PPlug}~\cite{liu2024llms+} aggregates historical interaction embeddings into a unified query-aware user representation.

\noindent\textbf{PHF Variants}:
To validate each level of the PHF hierarchy, we implement variants that operationalize habitus and field through different mechanisms.
(1) \textbf{T-H}~\cite{zhang2024guided} constructs habitus by summarizing user histories into textual persona descriptions using \textbf{Qwen2.5-14B-Instruct}.
(2) \textbf{T-F} constructs field similar with T-H to summarize collective preferences and behavioral tendencies within each user group.
(3) \textbf{L-F}~\cite{shi2025retrieval} constructs field through cross-user retrieval over behavior pools from similar users.
(4) \textbf{T-H+T-F} combines textual habitus summaries with text-based field descriptions.
(5) \textbf{T-H+L-F} combines textual habitus summaries with cross-user retrieval-based field modeling.

\begin{table*}[]
\centering
\resizebox{\textwidth}{!}{%

\begin{tabular}{llclcclcclccclccclccc}
\hline
\multirow{2}{*}{\textbf{Method}} &  & \textbf{LaMP-1}     &                      & \multicolumn{2}{c}{\textbf{LaMP-2}} &                      & \multicolumn{2}{c}{\textbf{LaMP-3}}      &                      & \multicolumn{2}{c}{\textbf{LaMP-4}}                 &                      & \multicolumn{2}{c}{\textbf{LaMP-5}}                 &                      & \multicolumn{2}{c}{\textbf{LaMP-7}}                 \\ \cline{3-3} \cline{5-6} \cline{8-9} \cline{11-12} \cline{14-15} \cline{17-18} 
                        &  & Acc$\uparrow$ &                      & Acc$\uparrow$  & F1$\uparrow$ &                      & MAE$\downarrow$ & RMSE$\downarrow$ &                      & R-1$\uparrow$ & R-L$\uparrow$ &                     & R-1$\uparrow$ & R-L$\uparrow$  &                      & R-1$\uparrow$ & R-L$\uparrow$  \\ \hline
$\mathrm{PHF}_{\text{Compass}}$                     &  & 0.631         & \multicolumn{1}{c}{} & 0.531          & 0.461        & \multicolumn{1}{c}{} & 0.260           & 0.579            & \multicolumn{1}{c}{} & 0.193         & 0.172             & \multicolumn{1}{c}{} & 0.516         & 0.453                & \multicolumn{1}{c}{} & 0.525         & 0.465             \\ \hline
w/o Habitus                &  & 0.501         &                      & 0.401         & 0.349        &                      & 0.321           & 0.649            &                      & 0.178         & 0.157                  &                      & 0.490         & 0.425         &                       & 0.491         & 0.431                \\ \hline
w/o Field               &  & 0.598        &                      & 0.527          & 0.435        &                      & 0.264           & 0.588            &                      & 0.199         & 0.176                &                      & 0.517         & 0.453         &                             & 0.529         & 0.467         &        \\ \hline
w/o Temporal            &  & 0.503         &                      & 0.538          & 0.471        &                      & 0.265           & 0.584            &                      & 0.201         & 0.180            &                      & 0.519         & 0.454                &                      & 0.518         & 0.460                \\ \hline
w/o Codebook            &  & 0.496         &                      & 0.508          & 0.428        &                      & 0.293           & 0.594            &                      & 0.156         & 0.138             &                      & 0.458         & 0.381              &                      & 0.500         & 0.439              \\ \hline
\end{tabular}
}
\caption{Ablation study of $\mathrm{PHF}_{\text{Compass}}$ on the LaMP benchmark.
“w/o” denotes removing the corresponding component from $\mathrm{PHF}_{\text{Compass}}$. 
“w/o Temporal” replaces temporally weighted aggregation with simple averaging, and “w/o Codebook” removes residual quantization and use original embedding.}
\label{tab:ablation}
\vspace{-10pt} 
\end{table*}

\subsection{Main Results}

The experimental results on the validation set are shown in Table~\ref{tab:main results}, and we can observe that:

(1)\textbf{ $\mathrm{PHF}_{\text{Compass}}$ consistently achieves the best performance across all tasks, yielding substantial improvements over the strongest flat-paradigm baselines on classification tasks and consistent gains on generation tasks.} Among other methods, retrieval-based approaches benefit from semantic matching (e.g., BM25 and Dense consistently outperform Recency and Zero-Shot), but remain limited by their reliance on individual behaviors. Aggregation-based methods such as PPlug and OPPU achieve stronger results by constructing holistic user representations, yet still underperform PHF variants. This suggests that while behavioral aggregation is beneficial, explicitly modeling the structure of user behavior is crucial for capturing the multi-level regularities that drive effective personalization.


(2) \textbf{Habitus and filed are complementary, and jointly modeling both yields stronger performance than modeling either level alone.} The results show that either level alone improves over flat baselines (e.g., T-H and T-F perform better than some flat paradigms on most tasks), and combining both yields further gains. Comparing different PHF implements, we can further find that $\mathrm{PHF}_{\text{Compass}}$ achieves strongest results overall, confirming our dedicated behavioral modeling outperforms text-based alternatives like T-H+T-F and T-H+L-F.

(3) \textbf{Different tasks benefit from different levels of the PHF hierarchy, highlighting the distinct roles of habitus and field in personalization.}.   Classification tasks (LaMP-1 to 3) primarily depend on user behavioral logic, so field modeling is particularly beneficial by compensating for sparse individual histories. As a result, field-enchanced methods like T-H, T-H+T-F and $\mathrm{PHF}_{\text{Compass}}$ obviously lead the classification tasks. In contrast, generation tasks rely on content stylistic consistency, so training-based aggregation methods like OPPU and PPlug remain competitive. $\mathrm{PHF}_{\text{Compass}}$, benefiting from the robust modeling of practice and habitus (see Sec.~\ref{app:ablation}), also show prominent performance on these tasks. Overall, $\mathrm{PHF}_{\text{Compass}}$ achieves the strongest performance across both task types, demonstrating the advantage of modeling multiple levels of user behavior.

\begin{table}[t!]
\resizebox{0.48\textwidth}{!}{%
\begin{tabular}{ccccccc}
\hline
\multirow{2}{*}{\begin{tabular}[c]{@{}c@{}}\textbf{Encoder} + \\ \textbf{Decoder}\end{tabular}} & \multicolumn{2}{c}{\textbf{LaMP-2}} &  & \multicolumn{2}{c}{\textbf{LaMP-4}} \\ \cline{2-3} \cline{5-6} 
                                                                                 & Acc           & F1            &  & R-1      & R-L       \\ \hline
Contriever+Qwen                                                                  & 0.531         & 0.461         &  & 0.199    & 0.176     \\ \hline
Contriever+LLaMA                                                                 & 0.510         & 0.423         &  & 0.215    & 0.193   \\ \hline
RoBERTa+Qwen                                                                     & 0.523         & 0.451         &  & 0.198    & 0.177    \\ \hline
RoBERTa+LLaMA                                                                    & 0.533         & 0.437         &  & 0.213    & 0.191    \\ \hline
\end{tabular}
}
\caption{Overall performance of models with different encoders and decoders on LaMP-2 and LaMP-4.}
\label{tab:generalizable}
\vspace{-10pt} 
\end{table}

\subsection{Ablation Study}\label{app:ablation}

We conduct an ablation study to evaluate the contribution of each component in $\mathrm{PHF}_{\text{Compass}}$. As shown in Table~\ref{tab:ablation}, we analyze the model from two perspectives:

\textbf{The effect of PHF constructs.} Removing habitus (\textit{w/o Habitus}) causes the largest performance degradation across all tasks, confirming that fine-grained individual behavioral representations are fundamental to effective personalization. Removing field (\textit{w/o Field}) leads to more moderate declines, particularly on classification tasks, indicating that collaborative signals are especially beneficial when behavioral reasoning rather than stylistic variation drives personalization. These results show that \textbf{both habitus and field contribute meaningfully to personalization, with habitus playing the dominant role.}

\textbf{The effect of implementation choices of} $\mathrm{PHF}_{\text{Compass}}$. Replacing temporal weighting (\textit{w/o Temporal}) with uniform averaging mainly affects tasks involving evolving user preferences, suggesting that temporal aggregation provides complementary gains beyond habitus construction itself. In addition, removing residual quantization (\textit{w/o Codebook}) consistently degrades performance across all tasks, with especially large drops on generation benchmarks. This highlights the importance of behavioral abstraction: without discrete practice representations, surface-level noise propagates through the entire hierarchy. Therefore, \textbf{the effectiveness of PHF also depends on our implementation choices}.


\begin{figure}[t]
    \centering
    \begin{subfigure}[b]{0.22\textwidth}
        \centering
        \includegraphics[width=\linewidth]{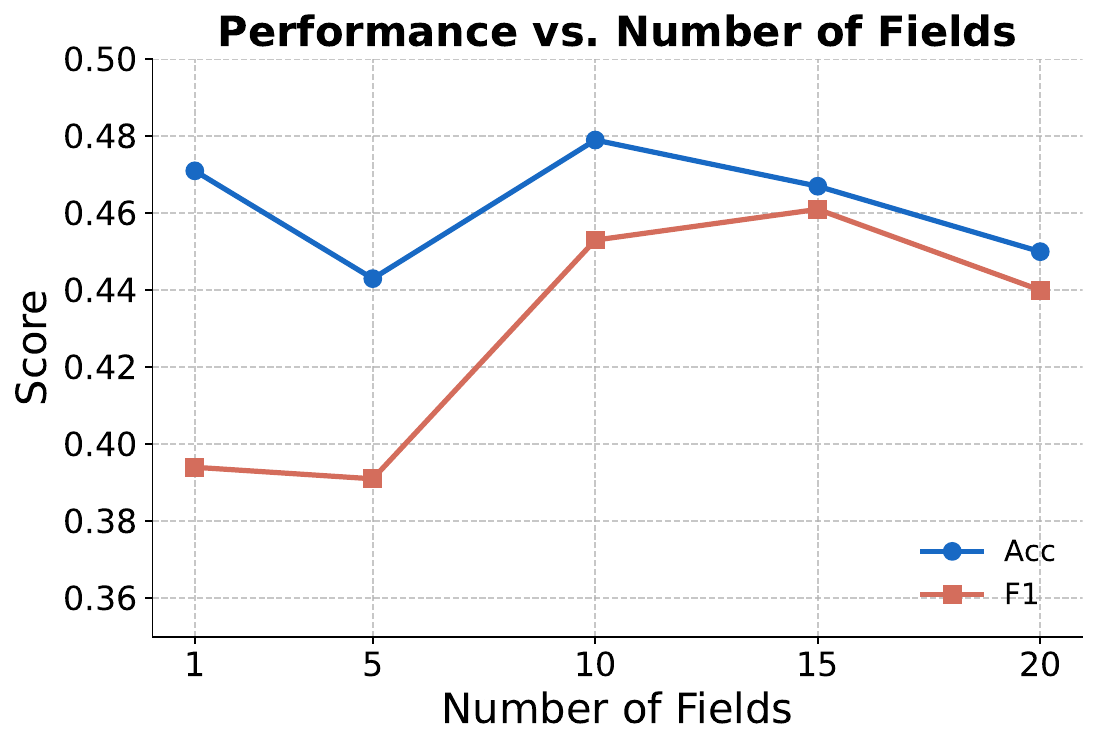}
        \caption{Performance of different field number on LaMP-2}
        \label{fig:num_field}
    \end{subfigure}
    \hfill
    \begin{subfigure}[b]{0.22\textwidth}
        \centering
        \includegraphics[width=\linewidth]{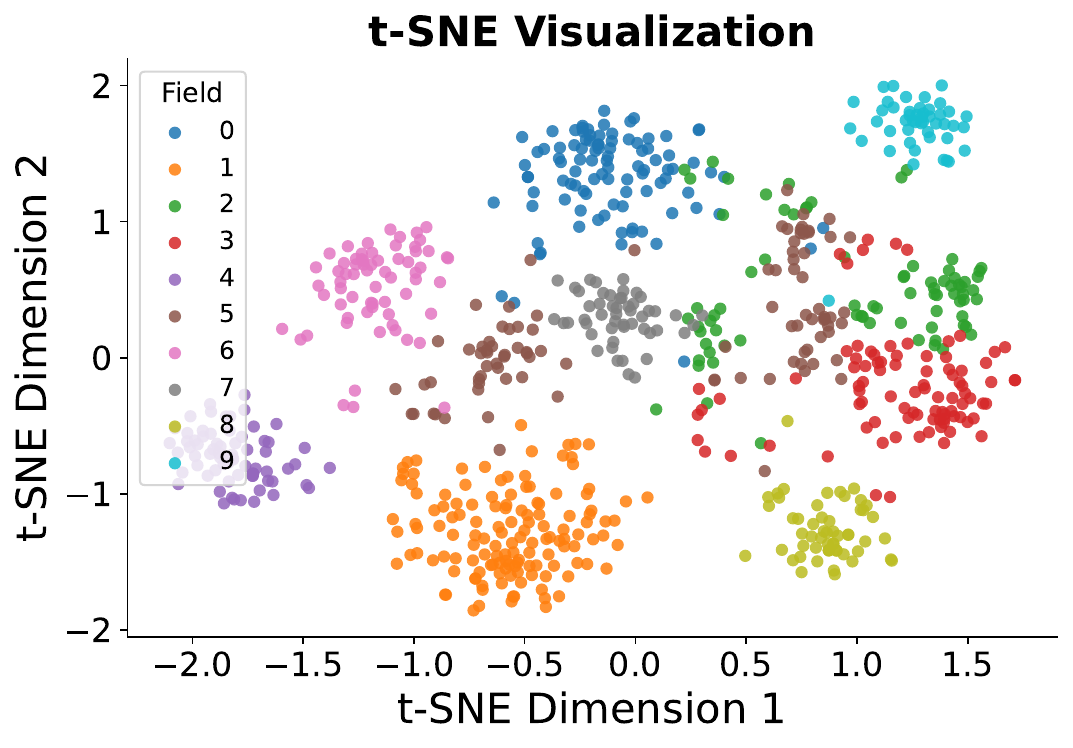}
        \caption{The t-SNE visualization of user cluster fields.}
        \label{fig:tsne}
    \end{subfigure}
    
    \caption{The influence of field granularity.}
    \label{fig:two_subfigures}
    \vspace{-10pt} 
\end{figure}
\subsection{Further Analysis}
We further conduct additional analysis on LaMP-2 (classification) and LaMP-4 (generation).



\subsubsection{Impact of Field Granularity}
A key design choice in PHF is the granularity of the field, which determines how finely the user population is partitioned into behavioral communities. This granularity is controlled by the number of clusters $K$: a smaller $K$ yields coarser fields that group diverse users together, potentially diluting shared behavioral signals with outlier patterns; a larger $K$ produces finer fields with higher internal homogeneity but fewer users per cluster, weakening the collaborative signal due to data sparsity. This trade-off is evident in Figure.~\ref{fig:num_field}: performance improves as $K$ increases from small values, but begins to decline once the fields become overly fragmented. The best balance is achieved at $K=10$, where both metrics are jointly optimized. Figure.~\ref{fig:tsne} further illustrates the corresponding t-SNE visualization of user habitus, showing that the induced fields are well separated and internally cohesive.


\begin{figure}[t]
\centering
\begin{subfigure}[b]{0.23\textwidth}
    \centering
    \includegraphics[width=\textwidth]{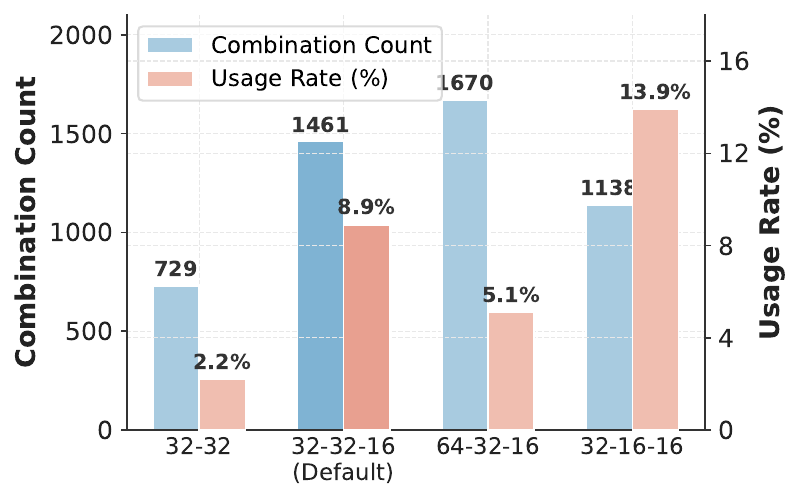}
    \caption{Codebook Size}
    \label{fig:codebook_size}
\end{subfigure}
\hfill
\begin{subfigure}[b]{0.23\textwidth}
    \centering
    \includegraphics[width=\textwidth]{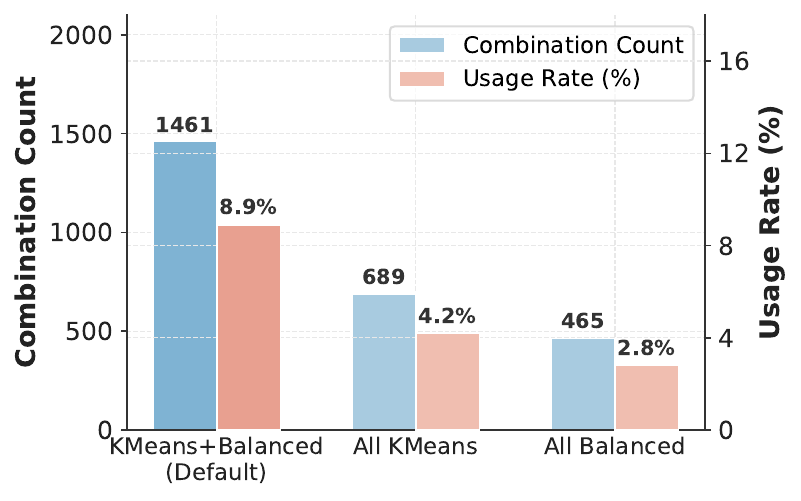}
    \caption{Initialization Strategy}
    \label{fig:codebook_init}
\end{subfigure}
\caption{Impact of codebook configuration. The notation $a$-$b$-$c$ indicates the codebook size for each layer. ``All KMeans'' applies standard K-Means initialization for all three layers.}
\label{fig:codebook}
\vspace{-15pt} 
\end{figure}

\subsubsection{Model-Agnostic Extensibility}


To validate the generalizability of $\mathrm{PHF}_{\text{Compass}}$ across different model architectures, we replace the default \textbf{Contriever} encoder and \textbf{Qwen2.5-7B-Instruct} decoder with \textbf{RoBERTa-large}~\cite{liu2019roberta} and \textbf{LLaMA-3-8B-Instruct}~\cite{grattafiori2024llama}, respectively. As shown in Table~\ref{tab:generalizable}, $\mathrm{PHF}_{\text{Compass}}$ maintains strong performance across all four encoder-decoder combinations, with only minor variation across configurations. This robustness demonstrates that the PHF framework is not tied to any specific backbone architecture. We further observe that LLaMA performs better on generation tasks, while Qwen achieves comparable or stronger results on classification tasks, suggesting that although backbone choice influences task preference, \textbf{the effectiveness of PHF remains consistent across architectures.}

\subsubsection{Impact of Codebook Configuration}

The codebook size directly determines the number of discrete practice categories and therefore the expressive capacity of user behavior modeling. To analyze this impact, we conduct experiments using different \textbf{codebook size}. As shown in Figure.~\ref{fig:codebook}(a), a small codebook fails to capture sufficient behavioral diversity, limiting personalization, while an excessively large codebook leads to sparse codeword usage and degraded performance due to under-trained entries. Furthermore, we analyze different \textbf{initialization strategy} (Figure.~\ref{fig:codebook}(b)). We find that the first two quantization layers primarily capture dominant behavioral patterns, for which standard K-Means provides a better fit than enforcing uniform code usage through balanced K-Means. In contrast, the third residual layer models small-magnitude residual signals and is more prone to code collapse under standard K-Means. This results in the final $32$-$32$-$16$ configuration with a hybrid initialization strategy: standard K-Means for the first two layers and balanced K-Means for the third. This achieves both high behavioral coverage and stable code utilization.




\subsubsection{Interpretability of $\mathrm{PHF}_{\text{Compass}}$}

$\mathrm{PHF}_{\text{Compass}}$ uses RQ-VAE to extract both shared patterns and personalized signals in a highly interpretable manner: (1) movies in user behavior histories with same semantic ID exhibit strong thematic coherence: e.g., \texttt{0-14-2} groups romance narratives set in fantastical worlds (\textit{Wonder Woman}, \textit{Howl’s Moving Castle}). (2) even under the same coarse tag like \textit{romance}, different level-1 codes reveal fine-grained distinctions: \texttt{0-} emphasizes otherworldly settings, while \texttt{16-} focuses on identity and emotional realism (\textit{Juno}, \textit{Safe Haven}). (3) identical movies yield distinct IDs when paired with different tags, e.g., \textit{Inception} as \textit{action} gives \texttt{8-21-12}, but as \textit{psychology} it becomes \texttt{15-21-12}, with only the level-1 code changing while deeper layers (\texttt{-21-12}) remain stable. These results suggest that the hierarchical quantization disentangles user-specific intent at the top level from shared semantic structure in deeper residual layers. More details are provided in Appendix~\ref{sec:inter}.





\section{Conclusion}
In this work, we propose PHF (Practice--Habitus--Field), a sociologically inspired framework that organizes LLM personalization into a unified behavioral hierarchy, jointly modeling within-user behavioral formation and cross-user behavioral regularity. Under this framework, we develop $\mathrm{PHF}_{\text{Compass}}$, a lightweight and model-agnostic implementation that abstracts user behaviors into denoised practices, aggregates them into stable habitus, and situates users within shared fields that capture collective behavioral regularities. Experiments on the LaMP benchmark demonstrate consistent improvements across both classification and generation tasks, while further analyses confirm the interpretability of the learned behavioral structures and the extensibility of the framework across different model architectures.

\section{Limitations}

This work has several limitations that suggest directions for future research.
 
Our experiments are conducted exclusively on the LaMP benchmark, which primarily consists of English-language text-based tasks with static user histories. The effectiveness of PHF on multilingual, multimodal, or domain-specific personalization scenarios (e.g., e-commerce, healthcare) remains to be validated. Additionally, since LaMP does not provide continuously evolving user interactions, we are unable to evaluate revolving habitus updates or dynamic field re-clustering in a streaming setting, which we consider an important direction for future work.
 
We evaluate personalization using standard classification and generation metrics such as Accuracy, F1, and ROUGE. These metrics may not fully capture the nuanced aspects of personalization quality, such as user satisfaction, perceived relevance, or behavioral consistency over time.
 
While $\mathrm{PHF}_{\text{Compass}}$ demonstrates one effective way to operationalize the PHF hierarchy, the specific implementations of each level are by no means the only options. For instance, habitus could be modeled through recurrent or attention-based temporal architectures rather than weighted aggregation, and fields could be induced through graph-based community detection or dynamic clustering rather than static K-Means. Exploring alternative implementations under the PHF framework is a promising avenue for future work.

\section{Ethical Considerations}

Personalization inherently involves modeling individual user behavior, which raises potential concerns regarding privacy and data usage. In this work, all experiments are conducted on the publicly available LaMP benchmark, which consists of anonymized user interaction data. Our framework operates without accessing other users' raw data during inference --- field embeddings are derived from cluster centroids learned during training, and no cross-user data sharing occurs at test time. Nevertheless, deploying personalization systems in real-world settings requires careful attention to user consent and data protection regulations. Additionally, clustering users into shared fields based on behavioral similarity may group users in ways that reinforce existing biases in the training data. We encourage future work to investigate fairness-aware extensions of the PHF framework and to evaluate personalization systems under broader ethical criteria beyond task performance.
 

\bibliography{custom}

\appendix

\section{Dataset Details}~\label{sec:data}
Detailed statistics and example input output pair for all six tasks are provided in Table~\ref{tab:data}. We use the time-split of the benchmark.

\begin{table*}[t!]
\centering
\resizebox{\textwidth}{!}{%
\begin{tabular}{ccccccccc}
\hline
\textbf{Task} & \textbf{\#Train} & \textbf{\#Dev} & \textbf{In Len} & \textbf{\begin{tabular}[c]{@{}c@{}}\#Cls or\\ Out Lens\end{tabular}} & \textbf{Hist Len} & \textbf{Input Format}                                                                                                                                                                                     & \textbf{\begin{tabular}[c]{@{}c@{}}Output \\ Format\end{tabular}}                           & \textbf{\begin{tabular}[c]{@{}c@{}}History \\ Format\end{tabular}}             \\ \hline
LaMP-1        & 6,542            & 1,500          & 51.43           & 2                                                                    & 84.15             & \begin{tabular}[c]{@{}c@{}}For an author who has written the paper with title ``\{title\}'', which reference \\ is related? Just answer with <1> or <2>. <1>: ``\{ref1\}'' <2>: ``\{ref2\}''\end{tabular} & <1>                                                                                         & \begin{tabular}[c]{@{}c@{}}title: \{title\} \\ abstract: \{abs\}\end{tabular}  \\ \hline
LaMP-2        & 5,073            & 1,410          & 92.39           & 15                                                                   & 86.76             & \begin{tabular}[c]{@{}c@{}}Which tag does this movie relate to among the following tags? Just \\ answer with the tag name. tags: [sci-fi, action, ...] description: \{movie\}\end{tabular}                & sci-fi                                                                                      & \begin{tabular}[c]{@{}c@{}}description: \{des\}\\  tag: \{tag\}\end{tabular}   \\ \hline
LaMP-3        & 20,000           & 2,500          & 128.18          & 5                                                                    & 185.40            & \begin{tabular}[c]{@{}c@{}}What is the score of the following review on a \\ scale of 1 to 5? Just answer with 1--5. review: \{review\}\end{tabular}                                                      & 3                                                                                           & \begin{tabular}[c]{@{}c@{}}text: \{review\} \\ score: \{score\}\end{tabular}   \\ \hline
LaMP-4        & 12,500           & 1,500          & 29.97           & 10.07                                                                & 204.59            & Generate a headline for the following article: \{article\}                                                                                                                                                & How I Got 'Rich'                                                                            & \begin{tabular}[c]{@{}c@{}}title: \{title\} \\ text: \{article\}\end{tabular}  \\ \hline
LaMP-5        & 14,682           & 1,500          & 162.34          & 9.71                                                                 & 87.88             & Generate a title for the following abstract of a paper: \{abstract\}                                                                                                                                      & \begin{tabular}[c]{@{}c@{}}Distributed Partial \\ Clustering\end{tabular}                   & \begin{tabular}[c]{@{}c@{}}title: \{title\} \\ text: \{abstract\}\end{tabular} \\ \hline
LaMP-7        & 13,437           & 1,498          & 29.72           & 16.96                                                                & 15.71             & Paraphrase the following tweet without any explanation: \{tweet\}                                                                                                                                         & \begin{tabular}[c]{@{}c@{}}gotta make the most of \\ my last full day in ktown\end{tabular} & text: \{tweet\}                                                                \\ \hline
\end{tabular}
}
\caption{Dataset statistics and input--output formats for different tasks in the LaMP benchmark. “\#” denotes the number of instances, “Cls” indicates classification tasks, and placeholders in braces \{\} are replaced with dataset-specific values.}
\label{tab:data}
\end{table*}

\section{Integration with RAG}

\begin{figure}
    \centering
    \begin{subfigure}[b]{0.23\textwidth}
        \centering
        \includegraphics[width=\linewidth]{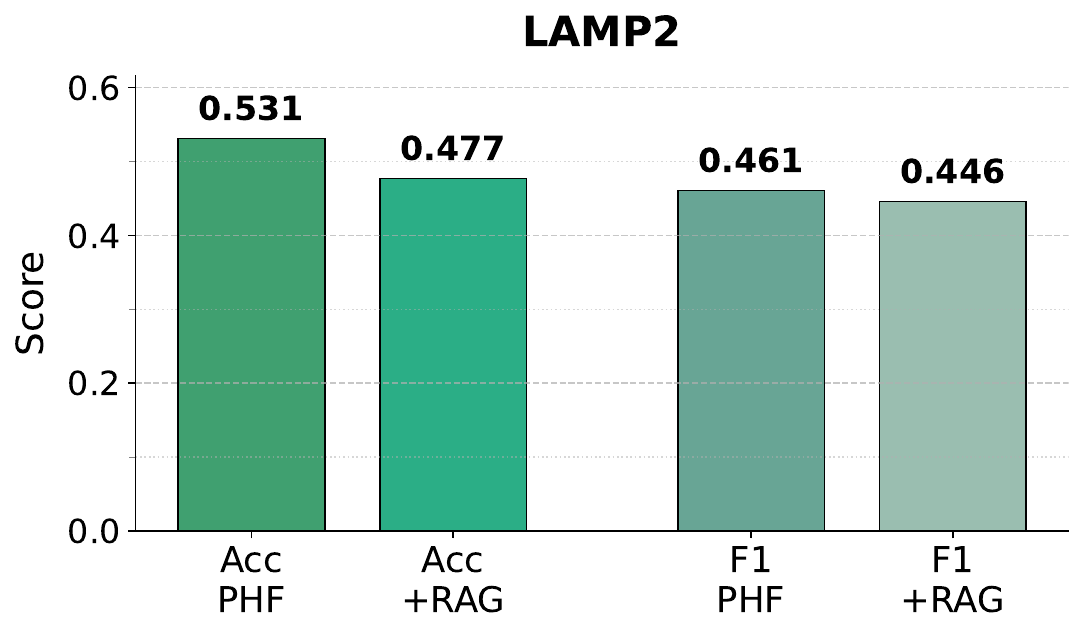}
        \caption{LaMP-2(C)}
        \label{fig:lamp2}
    \end{subfigure}
    \hfill
    \begin{subfigure}[b]{0.23\textwidth}
        \centering
        \includegraphics[width=\linewidth]{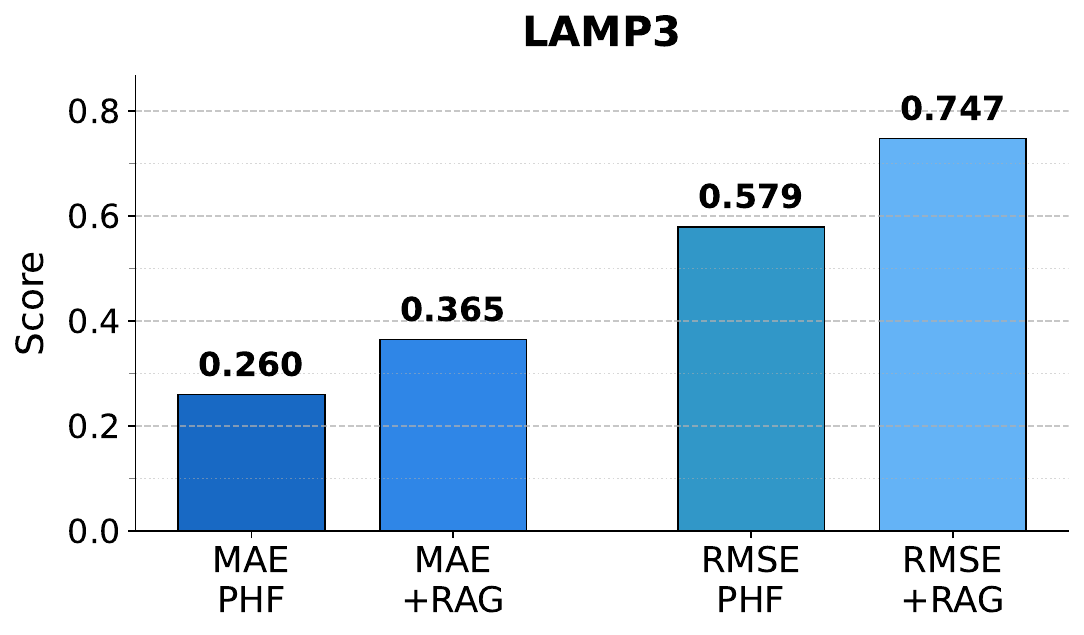}
        \caption{LaMP-3(C)}
        \label{fig:lamp3}
    \end{subfigure}
    
    \vspace{1em} 
    
    \begin{subfigure}[b]{0.23\textwidth}
        \centering
        \includegraphics[width=\linewidth]{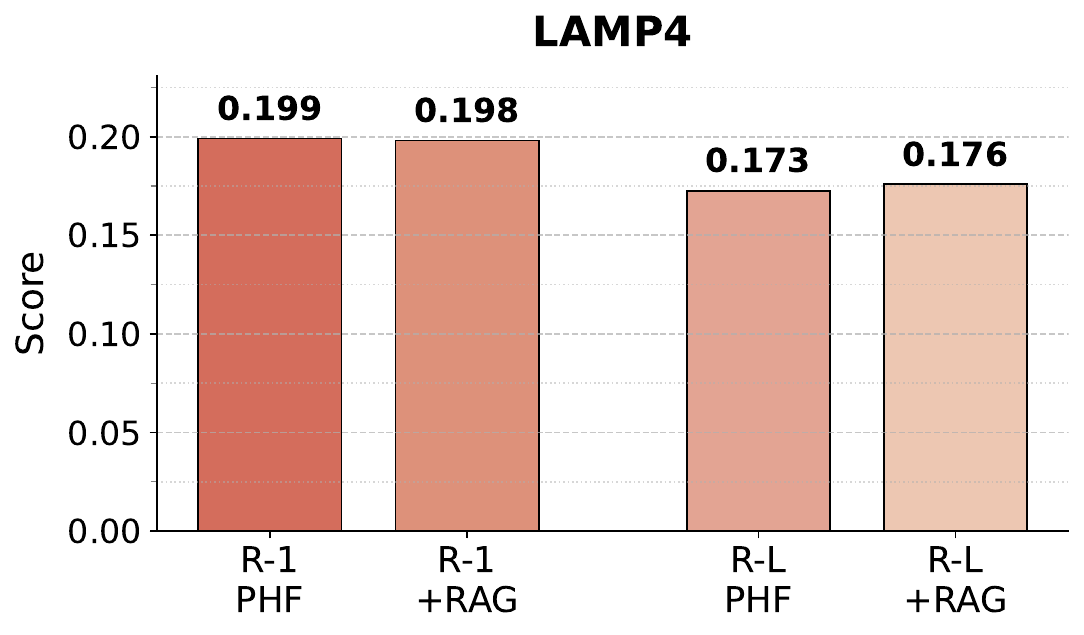}
        \caption{LaMP-4(G)}
        \label{fig:lamp4}
    \end{subfigure}
    \hfill
    \begin{subfigure}[b]{0.23\textwidth}
        \centering
        \includegraphics[width=\linewidth]{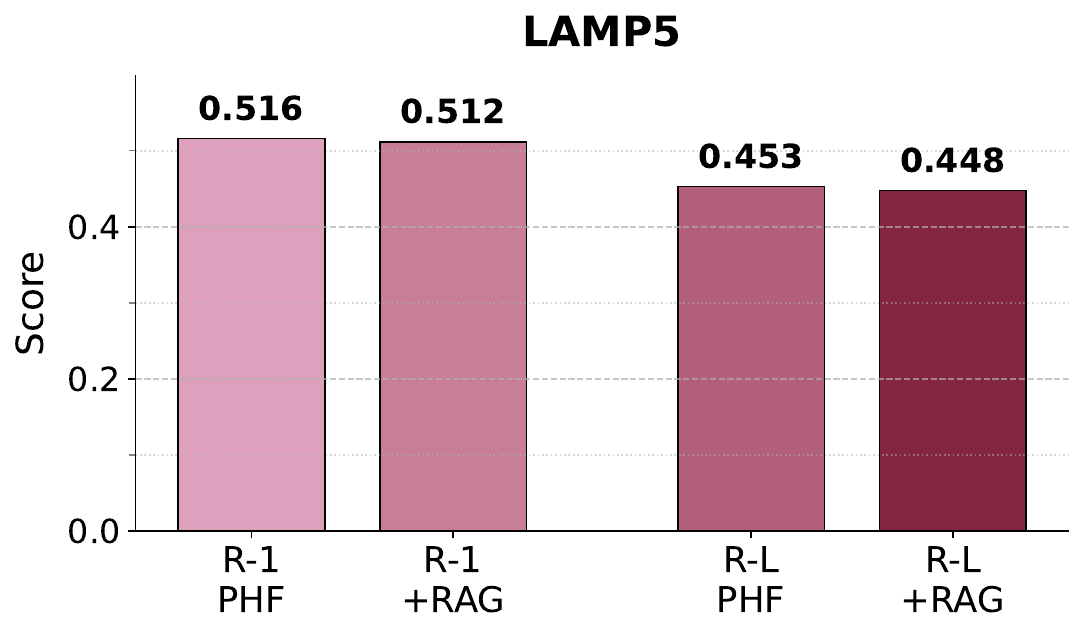}
        \caption{LaMP-5(G)}
        \label{fig:lamp5}
    \end{subfigure}
    
    \caption{Performance comparison across LaMP-2 to 5. Each subfigure shows scores for PHF and PHF+RAG variants.}
    \label{fig:lamp_all}
\end{figure}

Our preliminary experiments show that retrieval-augmented generation (RAG) can provide useful personalization signals. We therefore investigate whether integrating RAG with our habitus and field embeddings further improves performance. As shown in Figure.~\ref{fig:lamp_all}, we concatenate retrieved user history with the PHF embeddings as additional context. Surprisingly, this integration consistently degrades performance across tasks. We attribute this to the noise and excessive length introduced by raw retrieved text, which dilutes the focused signal from our compact latent embeddings. The effect is especially pronounced in classification tasks (LaMP-2 and LaMP-3), where performance drops significantly. In contrast, generation tasks (LaMP-4 and LaMP-5) show only minor fluctuations, suggesting greater robustness to input noise. These results indicate that our latent habitus and field embeddings already capture essential user characteristics effectively, eliminating the need for redundant or noisy textual context from RAG.

\section{Interpretability}~\label{sec:inter}

To further exhibit the interpretability of the Semantic ID based practice expression, we further exhibit some examples as shown in Table~\ref{tab:representative_movies_wide} and ~\ref{tab:same_movie_diff_tags}.

\begin{table*}[t]
\centering
\caption{Representative movies and their corresponding semantic IDs on LaMP-2}
\label{tab:representative_movies_wide}
\setlength{\tabcolsep}{6pt} 
\small
\begin{tabularx}{\textwidth}{@{}l l >{\raggedright\arraybackslash}X@{}}
\toprule
\textbf{Semantic ID} & \textbf{Genre} & \textbf{Movie (Title + Summary)} \\
\midrule

7-10-1 & fantasy &
\textit{Avengers: Infinity War} – Heroes unite to stop Thanos from collecting the six Infinity Stones and imposing his will on reality. \\
7-10-1 & fantasy &
\textit{The Lord of the Rings} – Aragorn defends Gondor while Frodo and Sam journey toward Mount Doom to destroy the One Ring. \\

8-10-1 & action &
\textit{Avengers: Endgame} – The remaining Avengers assemble one last time to reverse Thanos’ snap and restore the universe. \\
8-10-1 & action &
\textit{Iron Man 2} – Tony Stark battles government pressure, rival inventors, and his own failing health as Iron Man. \\

7-18-3 & fantasy &
\textit{Harry Potter} series – A young wizard uncovers secrets about his parents and battles dark forces throughout his school years. \\
7-18-3 & fantasy &
\textit{Star Wars: The Last Jedi} – Rey seeks training from a reluctant Luke Skywalker as the Resistance prepares for its final stand. \\

0-14-2 & romance &
\textit{Wonder Woman} – An Amazon princess leaves her island to fight in World War I and discover the power of love and truth. \\
0-14-2 & romance &
\textit{Howl’s Moving Castle} – A cursed girl finds refuge and romance in a wizard’s enchanted, walking castle. \\

16-14-2 & romance &
\textit{Juno} – A sharp-witted teenager navigates an unexpected pregnancy and makes an unconventional choice about her future. \\
16-14-2 & romance &
\textit{Safe Haven} – A mysterious woman hiding from her past finds healing and love in a small coastal town. \\

\bottomrule
\end{tabularx}
\caption{Semantic ID example for some combinations.}
\end{table*}

\begin{table}[t]
\centering
\small
\resizebox{\columnwidth}{!}{%
\setlength{\tabcolsep}{5pt}
\begin{tabular}{@{}p{4.8cm}p{1.6cm}l@{}}
\toprule
\textbf{Movie (Title + Summary)} & \textbf{Tag} & \textbf{Semantic ID} \\
\midrule

\multirow{3}{=}{\textit{Inception} – A thief who steals secrets from dreams is hired to implant an idea into a target’s subconscious.} 
& action & 8-21-12 \\
& sci-fi & 5-21-12 \\
& twist ending & 12-21-12 \\

\midrule

\multirow{3}{=}{\textit{Interstellar} – Explorers use a wormhole to travel across galaxies and save humanity from extinction.} 
& action & 26-26-7 \\
& dystopia & 14-26-7 \\
& twist ending & 12-26-7 \\

\midrule

\multirow{3}{=}{\textit{The Dark Knight} – Batman battles the Joker, a criminal who tests the moral limits of Gotham’s heroes.} 
& action & 21-25-13 \\
& dystopia & 14-25-13 \\
& violence & 20-25-13 \\
\bottomrule
\end{tabular}
}
\caption{Same movie paired with diverse user tags on LaMP-2}
\label{tab:same_movie_diff_tags}
\end{table}

\section{Detailed Procedure for Encoder Refinement}~\label{sec:CL}

Although the main framework only requires a lightweight interaction encoder, in practice we further refine the encoder to better align contextual semantics with user-specific behavioral expression. Specifically, given an interaction
\[
h_{it} = (q_{it}, r_{it}),
\]
we separately encode the query and response:
\begin{equation}
\mathbf{q}_{it}
=
\mathrm{Enc}(q_{it}),
\qquad
\mathbf{r}_{it}
=
\mathrm{Enc}(r_{it}).
\end{equation}

The final interaction embedding is constructed through weighted fusion:
\begin{equation}
\mathbf{x}_{it}
=
\mathrm{norm}
\left(
\alpha \cdot \mathrm{norm}(\mathbf{q}_{it})
+
\beta \cdot \mathrm{norm}(\mathbf{r}_{it})
\right),
\end{equation}
where $\alpha$ and $\beta$ balance contextual information and user-specific expression.

To further improve semantic consistency, we introduce a contrastive alignment objective between two complementary semantic views. The first view directly encodes the concatenated interaction:
\begin{equation}
\mathbf{x}_{it}^{(1)}
=
\mathrm{norm}
\left(
\mathrm{Enc}
(
[q_{it}, r_{it}]
)
\right),
\end{equation}
while the second view corresponds to the weighted fusion representation:
\begin{equation}
\mathbf{x}_{it}^{(2)}
=
\mathrm{norm}
\left(
\alpha \cdot \mathrm{norm}(\mathbf{q}_{it})
+
\beta \cdot \mathrm{norm}(\mathbf{r}_{it})
\right).
\end{equation}

We treat the two representations as a positive pair and optimize a symmetric InfoNCE objective:
\begin{equation}
\begin{aligned}
\mathcal{L}_{\text{cont}}
=
\frac{1}{2}
\Big[
&
\mathcal{H}
\left(
\frac{
\mathrm{sim}
(
\mathbf{x}_{it}^{(1)},
\mathbf{x}_{it}^{(2)}
)
}{\tau}
\right)
\\
+
&
\mathcal{H}
\left(
\frac{
\mathrm{sim}
(
\mathbf{x}_{it}^{(2)},
\mathbf{x}_{it}^{(1)}
)
}{\tau}
\right)
\Big],
\end{aligned}
\end{equation}
where $\mathrm{sim}(\cdot,\cdot)$ denotes cosine similarity, $\tau$ is a temperature parameter, and $\mathcal{H}$ denotes cross-entropy loss.

This refinement objective encourages the fused representation to preserve the joint semantics captured by direct interaction encoding while maintaining a disentangled representation of contextual grounding and user-specific behavioral expression.

\begin{algorithm}[t]
\caption{Encoder Refinement with Contrastive Alignment}
\label{alg:encoder}
\small
\begin{algorithmic}[1]
\REQUIRE Interaction set $\mathcal{H}$
\FOR{each interaction $h_{it}=(q_{it},r_{it})$}
    \STATE Encode query embedding $\mathbf{q}_{it}$
    \STATE Encode response embedding $\mathbf{r}_{it}$
    \STATE Construct fused representation $\mathbf{x}_{it}^{(2)}$
    \STATE Encode concatenated interaction $\mathbf{x}_{it}^{(1)}$
    \STATE Compute contrastive alignment loss
    \STATE Update encoder parameters
\ENDFOR
\RETURN Refined encoder $\mathrm{Enc}(\cdot)$
\end{algorithmic}
\end{algorithm}

\section{Detailed Procedure for Balanced K-Means Initialization}~\label{sec:bkl}

To improve codebook stability during residual quantization, especially in deeper quantization layers where residual magnitudes become progressively smaller, we initialize the final codebook using balanced K-Means.

Given residual representations
\[
\mathcal{R}
=
\{
\mathbf{r}_1,
\dots,
\mathbf{r}_N
\},
\]
balanced K-Means aims to partition residuals into $K$ approximately balanced clusters:
\begin{equation}
\min_{\{\mathcal{C}_k\}}
\sum_{k=1}^{K}
\sum_{\mathbf{r}\in\mathcal{C}_k}
\|
\mathbf{r}-\boldsymbol{\mu}_k
\|_2^2,
\end{equation}
subject to:
\begin{equation}
|\mathcal{C}_k|
\approx
\frac{N}{K},
\qquad
\forall k.
\end{equation}

Here,
$\boldsymbol{\mu}_k$
denotes the centroid of cluster
$\mathcal{C}_k$.
Compared with standard K-Means, balanced clustering prevents excessive concentration of residuals into a small number of centroids, thereby improving codebook utilization and stabilizing subsequent quantization training.

In practice, we only apply balanced K-Means to initialize the deepest residual codebook, while shallower codebooks are initialized using standard K-Means.

\begin{algorithm}[t]
\caption{Balanced K-Means Initialization}
\label{alg:balanced_kmeans}
\small
\begin{algorithmic}[1]
\REQUIRE Residual set $\mathcal{R}$, number of clusters $K$
\STATE Initialize centroids $\{\boldsymbol{\mu}_k\}_{k=1}^{K}$
\REPEAT
    \STATE Assign residuals to nearest centroids under balance constraints
    \STATE Update centroids using assigned residuals
\UNTIL{cluster assignments converge}
\RETURN Balanced codebook centroids $\{\boldsymbol{\mu}_k\}_{k=1}^{K}$
\end{algorithmic}
\end{algorithm}

\begin{figure}[t]
    \centering
    \begin{subfigure}[b]{0.23\textwidth}
        \centering
        \includegraphics[width=\linewidth]{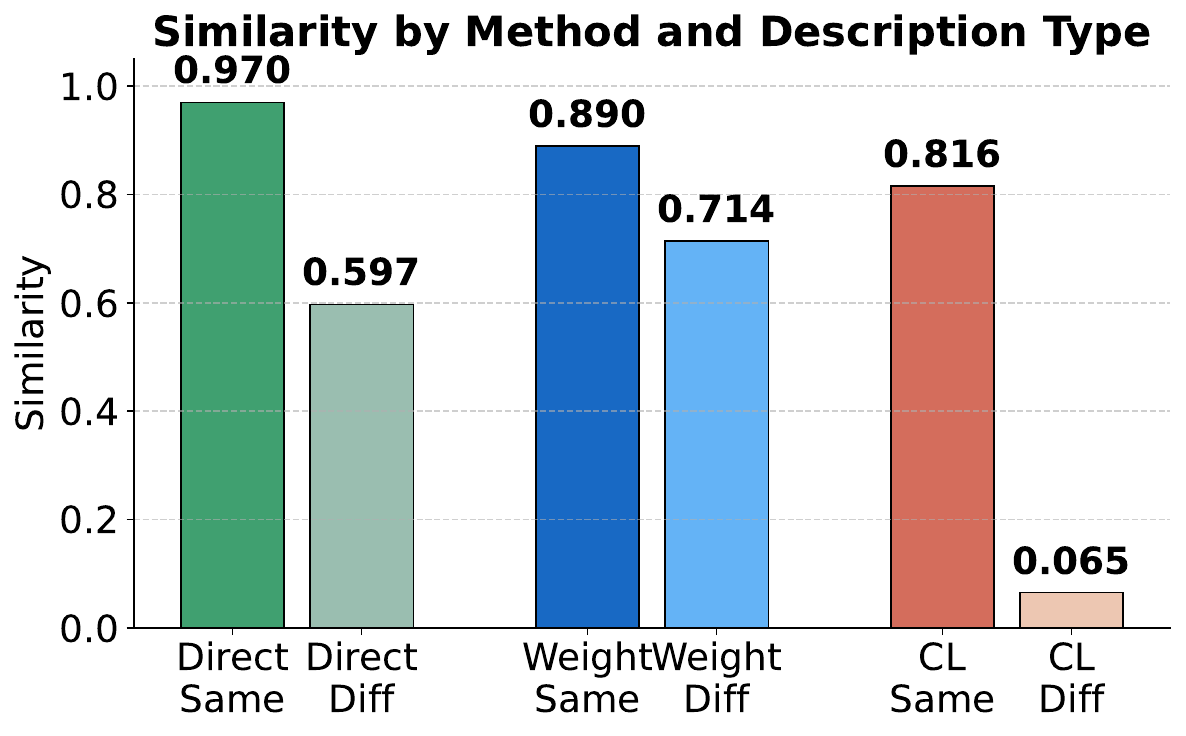}
        \caption{Mean similarity of different encoding methods.}
        \label{fig:sim}
    \end{subfigure}
    \hfill
    \begin{subfigure}[b]{0.23\textwidth}
        \centering
        \includegraphics[width=\linewidth]{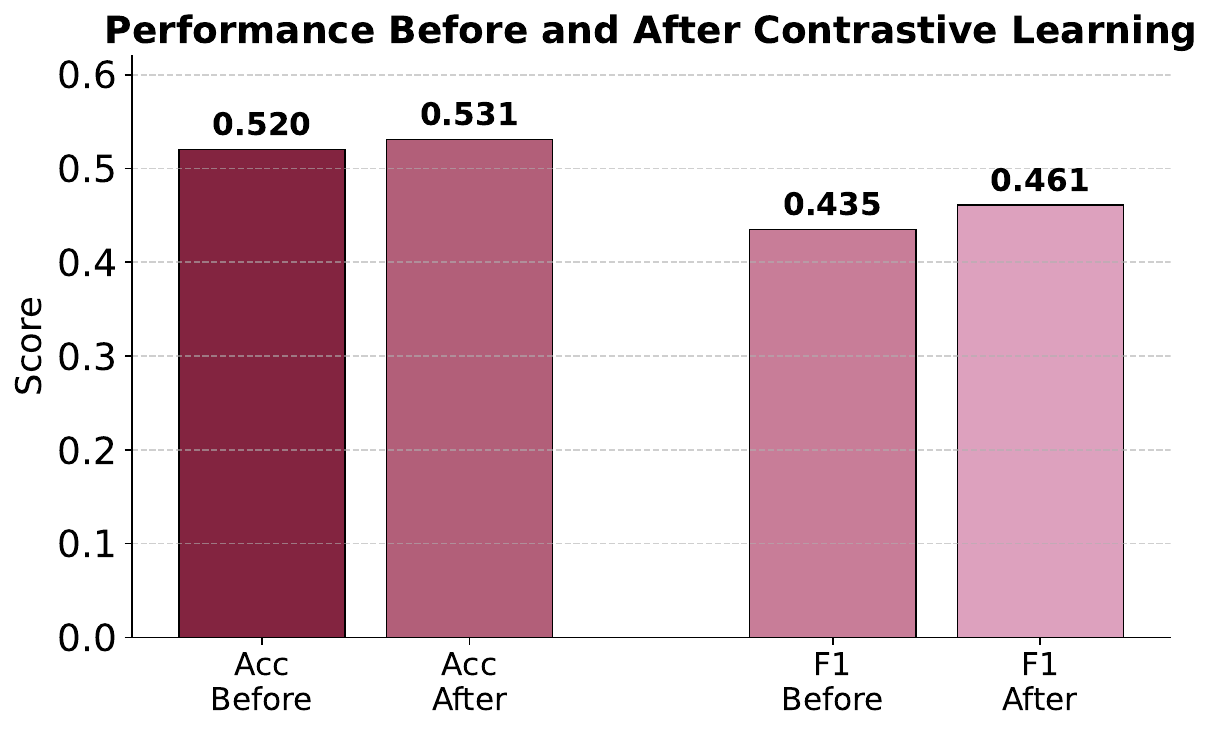}
        \caption{Performance with and without CL on LaMP-2.}
        \label{fig:CL}
    \end{subfigure}
    
    \caption{The impact of contrastive learning on the encoder.}
    \label{fig:CL}
\end{figure}

\section{Impact of Contrastive Learning on the Encoder}

Contrastive learning effectively reduces the spurious similarity caused by length imbalance between query and response, and avoids the artificial similarity introduced by simple weighted summation. Taking LaMP-2 as an example shown in Fig~\ref{fig:CL}, where the query is a long movie description and the response is only 1–2 words, we observe that directly encoding the same query with different responses yields nearly identical embeddings (similarity = 0.970), making them hard to distinguish. While weighted fusion improves separability, it introduces extra similarity: the same response paired with different queries shows increased similarity (from 0.597 to 0.714), which is still suboptimal. In contrast, our contrastive learning approach shown in Fig~\ref{fig:CL}(a) achieves the best separation by aligning the fused representation with its reconstructed counterpart which greatly helped the downstream RQ-VAE. Fig~\ref{fig:CL}(b) further validates its effectiveness: removing contrastive learning leads to clear drop in both Accuracy and F1.

\section{Implementation Details}~\label{sec:detail}

We use Contriever~\cite{izacard2021unsupervised} as the default interaction encoder and Qwen2.5-7B-Instruct~\cite{qwen2.5} as the decoder. In the extensibility experiments (Section~\ref{tab:generalizable}), we additionally evaluate RoBERTa-large~\cite{liu2019roberta} as the encoder and LLaMA-3-8B-Instruct~\cite{grattafiori2024llama} as the decoder.

The codebook consists of $L = 3$ layers with codebook sizes of $32$, $32$, and $16$ for layers 1, 2, and 3, respectively. Codebooks in layers 1 and 2 are initialized using standard K-Means on the training data, while layer 3 employs balanced K-Means initialization to promote uniform codeword usage. The commitment loss coefficient $\lambda_v$ and the codebook update coefficient are both set to $0.25$. The usage regularization weight $\lambda_u$ in Equation~(8) is set to $0.05$. The temperature is set to $0.05$. The random seed we use are $0,1,42$. We use rouge\_score package to calculate ROUGE.

The hyperparameters $\alpha$ and $\beta$ for contrastive learning are set to $1.0$ and $1.2$, respectively. The default number of retrieved histories is $k = 8$. For field-aware user clustering, users are partitioned into $K = 10$ groups by default. We conduct contrastive learning for all tasks except LaMP-1 and LaMP-7 due to data format constraints.

The MLP consists of two layers with dimensions $\text{encoder\_dim} \rightarrow \text{decoder\_dim} \rightarrow \text{decoder\_dim}$, using GELU activations. We optimize the model using AdamW with a batch size of $4$ per GPU and a learning rate of $1 \times 10^{-4}$. Gradient accumulation with step size $8$ is applied to simulate the target effective batch size. All experiments are conducted on 8 NVIDIA A100 (80GB) GPUs.

\section{Licenses}
We use publicly available datasets and models that are already licensed. Our code will be released if accepted with MIT License.

\end{document}